\pdfoutput=1

\documentclass[11pt]{article}

\usepackage[final]{latex/acl}
\usepackage{amsmath}
\usepackage{amsfonts}
\usepackage[normalem]{ulem}
\usepackage{times}
\usepackage{booktabs}
\usepackage{latexsym}
\usepackage{multirow}
\usepackage{graphicx}
\usepackage{subcaption}
\usepackage{float}

\usepackage[T1]{fontenc}
\usepackage{enumitem}

\newcommand{\name}{{HALP}}

\usepackage[utf8]{inputenc}

\usepackage{microtype}

\usepackage{inconsolata}
\usepackage{natbib}                %
\usepackage{graphicx}
\usepackage{float}
\usepackage{tabularx}  %
\usepackage{array}     %
\newcolumntype{L}[1]{>{\raggedright\arraybackslash}p{#1}} %
\newcolumntype{C}[1]{>{\centering\arraybackslash}p{#1}}   %

\title{{\name}: Detecting Hallucinations in Vision-Language Models\\without Generating a Single Token}

\author{
  Sai Akhil Kogilathota\textsuperscript{1} \quad
  Sripadha Vallabha E G\textsuperscript{1} \quad
  Luzhe Sun\textsuperscript{2} \qquad
  Jiawei Zhou\textsuperscript{1} \\
  \textsuperscript{1}Stony Brook University \quad
  \textsuperscript{2}Toyota Technological Institute at Chicago \\
  \texttt{luzhesun@ttic.edu} \\
  \texttt{\{saiakhil.kogilathota,\, sripadhavallab.eg,\, jiawei.zhou.1\}\;@stonybrook.edu}
}

\begin{document}
\maketitle

\begin{abstract}

Hallucinations remain a persistent challenge for vision–language models (VLMs), which often describe nonexistent objects or fabricate facts. Existing detection methods typically operate \emph{after} text generation, making intervention both costly and untimely. We investigate whether hallucination risk can instead be predicted \emph{before} any token is generated by probing a model’s internal representations in a single forward pass.
Across a diverse set of vision–language tasks and eight modern VLMs, including Llama-3.2-Vision, Gemma-3, Phi-4-VL, and Qwen2.5-VL, we examine three families of internal representations: (i) visual-only features without multimodal fusion, (ii) vision-token representations within the text decoder, and (iii) query-token representations that integrate visual and textual information before generation.
Probes trained on these representations achieve strong hallucination-detection performance without decoding, reaching up to 0.93 AUROC on Gemma-3-12B, Phi-4-VL 5.6B, and Molmo 7B. Late query-token states are the most predictive for most models, while visual or mid-layer features dominate in a few architectures (e.g., $\sim$0.79 AUROC for Qwen2.5-VL-7B using visual-only features). These results demonstrate that (1) hallucination risk is detectable \emph{pre-generation}, (2) the most informative layer and modality vary across architectures, and (3) lightweight probes has the potential to enable early abstention, selective routing, and adaptive decoding to improve both safety and efficiency.

\end{abstract}

\section{Introduction}

Recent vision–language models (VLMs) such as Gemma-3 \citep{steiner2024paligemma2familyversatile}, LLaVA-Next \citep{liu2024llavanext}, Llama-3.2-Vision \citep{meta2024llama32vision}, 
Phi-4-VL \citep{li2022blip}, Molmo \citep{molmo2023}, Qwen2.5-VL \citep{bai2024qwen25vl}, SmolVLM \citep{smolvlm2024}, and FastVLM \citep{fastvlm2024} have achieved remarkable progress in generating coherent, human-like text grounded in visual inputs, redefining the frontier of multimodal AI \citep{hurst2024gpt, li20254d, li-etal-2025-text}. Yet despite these advances, VLMs remain prone to hallucination, producing descriptions of nonexistent objects, invented attributes, or confident but unsupported claims that are factually inconsistent with the image. Such failures undermine the reliability and trustworthiness of these systems, especially in high-stakes applications such as autonomous navigation, medical imaging, and assistive technologies \citep{rohrbach-etal-2018-object, li-etal-2023-evaluating}.

Current approaches to mitigating hallucinations in vision–language models are largely \textbf{reactive}. Post hoc evaluation metrics such as CHAIR \citep{rohrbach-etal-2018-object}, POPE \citep{li2023pope}, and FaithScore \citep{jing2024faithscore} require complete caption generation to identify hallucinated content, making them computationally expensive and unsuitable for real-time assessment. Other methods aim to reduce hallucinations \textit{during} generation through decoding-time interventions such as HALC \citep{chen2024halcobjecthallucinationreduction}, Uncertainty-Guided Dropout Decoding \citep{fang2024uncertaintytrustenhancingreliability}, and perception-driven grounding augmentation \citep{ghosh2025visualdescriptiongroundingreduces}. However, these techniques still depend on autoregressive decoding and cannot estimate hallucination risk before generation begins. This reveals a critical gap: few existing methods exploit internal model representations to anticipate hallucinations preemptively and enable real-time risk detection prior to decoding.

\begin{figure*}
    \centering
    \includegraphics[width=1.0\linewidth]{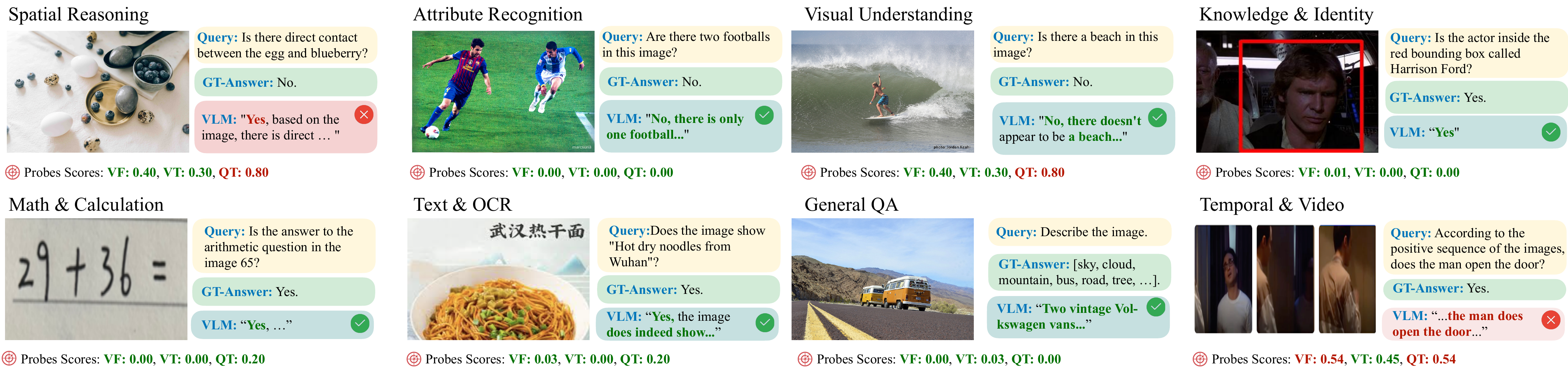}
    \caption{Pre-generation hallucination prediction scores on examples across 8 different VLM task domains (benchmarking data distribution in Fig.~\ref{fig:Dataset}(a)) for Qwen2.5-VL-7B. The probes scores reflect how likely the probe thinks the model \textit{will  hallucinate} based on different \textit{pre-generation features} (VF, VT, QT as defined in Sec.~\ref{subsec:feature}). 
    }\
    \vspace{-20pt}
    \label{fig:domain-type}
\end{figure*}

To address this gap, we introduce \textbf{\name} (\textit{HALlucination Prediction via Pre-Generation Probing}), a lightweight framework that predicts hallucination risk from \textit{pre-generative} internal states of vision–language models. {\name} performs a \textbf{single forward pass} over an image–query pair and extracts three types of representations: (1) pooled visual features from the vision encoder, (2) decoder states at the last vision tokens across multiple layers, and (3) decoder states at the final text query token at corresponding depths. Simple MLP probes trained on these representations estimate whether a model is likely to hallucinate based on the visual context \textit{before} any token is generated. Once trained, these probes require neither hallucination labels nor decoding, enabling efficient, generation-free risk prediction.

To assess generality, we construct a 10k-example benchmark covering diverse hallucination stressors, including object presence (AMBER \citep{an2023amber}, POPE \citep{li2023pope}), diagram and math reasoning (MathVista \citep{lu2023mathvista}), compositional and open-ended knowledge (MME \citep{fu2023mme}), illusion-style traps (HaloQuest \citep{cheng2023haloquest}), and targeted hallucination cases (HallusionBench \citep{guan2023hallusionbench}).
Figure~\ref{fig:domain-type} shows qualitative hallucination-detection examples across our evaluation domains, where {\name}’s pre-generation probes estimate hallucination risk before any output tokens are generated. 
High scores predicts VLM hallucinations, while low scores align with truthful responses.
Experiments on eight state-of-the-art open-source VLMs show that hallucination signals are accessible \textit{before generation}: 
probes trained on internal states reach high AUROC without decoding, with query-token representations near the top of the decoder most predictive ($\sim$0.93 for Gemma-3, Phi-4-VL, and Molmo). The most informative locus varies by architecture while deep query-token states dominate for most models, others such as Qwen-2.5-VL and Llama-3.2-Vision achieve strong detection from only visual features ($\sim$0.79 and $\sim$0.77, repsectively) or mid-decoder vision tokens, suggesting distinct failure pathways across models. Overall, lightweight probes generalize across architectures and representation types, offering a practical, low-cost solution for early hallucination risk assessment.\footnote{Code \& data at \url{https://github.com/Zesearch/HALP}.}

\section{Background and Related Work}

\paragraph{Hallucination in Vision-Language Models}
Hallucination in vision–language models (VLMs) refers to the generation of content that is inconsistent with the visual input. Prior work has identified multiple forms of hallucination, including object hallucinations, attribute hallucinations, and relational hallucinations \citep{rohrbach-etal-2018-object, song2025head}. Such errors undermine the reliability of VLMs, particularly in safety-critical settings such as medical imaging, autonomous navigation, and assistive technologies \citep{rohrbach-etal-2018-object, li-etal-2023-evaluating, jiang2025robustifying}. Hallucinations are commonly attributed to misalignment between linguistic priors and visual grounding, as well as biases introduced during training. For instance, models may describe nonexistent objects, assign incorrect attributes to visible entities, or infer spurious relationships between objects. These failure modes highlight a persistent disconnect between linguistic fluency and visual grounding, underscoring the need for effective hallucination detection and mitigation in vision–language systems.

\paragraph{Representation Probing}
Representation probing is a diagnostic approach where lightweight classifiers or regressors are trained on internal model activations to assess the encoding of specific properties. In NLP, probes have been used to capture syntactic dependencies, part-of-speech tags, and coreference relations \citep{hewitt-liang-2019-designing,marvin-linzen-2018-targeted, gottesman-geva-2024-estimating}. In vision, linear probes on convolutional features have been employed to reveal emergent object detectors in scene-classification networks \citep{zhou2015objectdetectorsemergedeep}. Recent work has extended probing to Vision Transformers, where intermediate layers encode rich class-specific and scene-level semantics \citep{chen2022adaptformeradaptingvisiontransformers, sheta-etal-2025-behavioral}. These studies demonstrate the effectiveness of probing as a lightweight tool for detecting task-relevant features \citep{sheta-etal-2025-behavioral}, motivating its use to identify hallucination signals early in VLMs.

\paragraph{Pre-Generation Hallucination Detection}
Recent advances have demonstrated that hallucination-related signals are present in the internal states of language models before generation occurs. \citet{orgad2024llmsknowmore} showed that LLMs' internal representations encode substantial information about truthfulness, with critical signals concentrated around specific tokens. They trained classifiers on hidden states to predict factual correctness, achieving high accuracy across multiple layers. Similarly, \citet{wang2023androids} designed probes trained on Transformer internal representations to predict hallucinatory behavior across grounded generation tasks, finding that probes can reliably detect hallucinations at many transformer layers, achieving 95\% of peak performance as early as layer 4. \citet{kadavath2022lmsknow} demonstrated that language models possess significant self-knowledge about their outputs through calibration experiments, supporting the hypothesis that error signals are accessible before generation completes. Building upon these insights, {\name} extends probing-based approaches to vision-language models, extracting internal representations from various stages of the VLM pipeline to predict hallucination risk before any tokens are generated.

\paragraph{Post-Generation Hallucination Detection and Mitigation}
Complementary to pre-generation approaches, substantial research has focused on detecting and mitigating hallucinations after content generation. Post-hoc detection methods include uncertainty-guided dropout decoding \citep{fang2024uncertaintytrustenhancingreliability}, adaptive focal-contrast decoding (HALC) \citep{chen2024halcobjecthallucinationreduction}, and perception-driven grounding augmentation \citep{ghosh2025visualdescriptiongroundingreduces}, among others. \citet{wan2024fava} introduced fine-grained hallucination detection with detailed error taxonomies, enabling precise identification of entity, relation, and contradictory errors in generated text. While these methods achieve high detection accuracy, they require complete sequence generation, making them computationally expensive and unsuitable for real-time applications where early intervention is crucial. Evaluation typically relies on metrics like CHAIR \citep{rohrbach-etal-2018-object}, which necessitates full caption generation for assessment.

\section{{\name}: Hallucination Prediction via Pre-Generation Probing}\label{sec: method}

\subsection{Vision-Language Model Preliminaries}

While there are variations, modern vision-language models (VLMs) generally adopt a three-component architecture \citep{liu2023visualinstructiontuning}. Given an input image $I$ and a text query $Q$, the processing pipeline operates as follows:

\noindent\textbf{Vision Encoding:} A vision encoder $\text{Enc}(\cdot)$, such as the one from CLIP \citep{radford2021learningtransferablevisualmodels}, processes the input image $I$, producing a sequence of vectors of \emph{visual features} $U =\{\mathbf{u}_1, \mathbf{u}_2, \ldots, \mathbf{u}_M\}, \mathbf{u}_i \in \mathbb{R}^{d_I}$, where $M$ is the number of visual patches and $d_I$ is the dimension of the feature.

\noindent\textbf{Multimodal Projection:} A vision-to-text projection layer $\mathcal{M}$ maps the visual features into the language model's input embedding space of dimension $d$, producing a sequence of \emph{vision tokens} $V=\{\mathbf{v}_1, \mathbf{v}_2, \ldots, \mathbf{v}_M\}$, with $\mathcal{M}(\mathbf{u}_i) \mapsto \mathbf{v}_i, \mathbf{v}_i\in \mathbb{R}^{d}$, enabling cross-modal fusion.

\noindent\textbf{Language Decoding:} The input text instruction (query) $Q$ is tokenized and embedded as
$X=\{\mathbf{x}_1, \mathbf{x}_2, \ldots, \mathbf{x}_N\}$, where each token embedding satisfies $\mathbf{x}_j \in \mathbb{R}^{d}$.
The concatenation of vision and text tokens yields the decoder input sequence
$S = [V; X] = \{\mathbf{s}_1, \ldots, \mathbf{s}_{M+N}\}$ with $\mathbf{s}_k \in \mathbb{R}^{d}$.
This sequence is processed by a Transformer-based decoder \citep{vaswani2017attention} with $L$ layers.
Each layer $\ell \in \{1,\ldots,L\}$ produces hidden states $\mathbf{h}^{(\ell)}_k \in \mathbb{R}^{d}$ for each token position $k \in \{1,\ldots,M+N\}$. The model then autoregressively generates the output text sequence $Y = \{ y_1, y_2, \ldots, y_T\}$  that is supposed to ground the vision information.

\paragraph{Overview}
Our goal is to explore whether VLMs already encode information about their knowledge certainty or propensity to hallucinate within their internal representations, before any text generation occurs. {\name} intercepts and analyzes the internal representations at three critical stages of this pipeline, extracting features that encode both visual understanding and multimodal reasoning. We hypothesize that these pre-generation features contain sufficient information to predict whether the model will produce hallucinated content, enabling real-time risk assessment without requiring complete sequence generation.

\begin{figure}[t]
    \centering
    \includegraphics[width=\columnwidth]{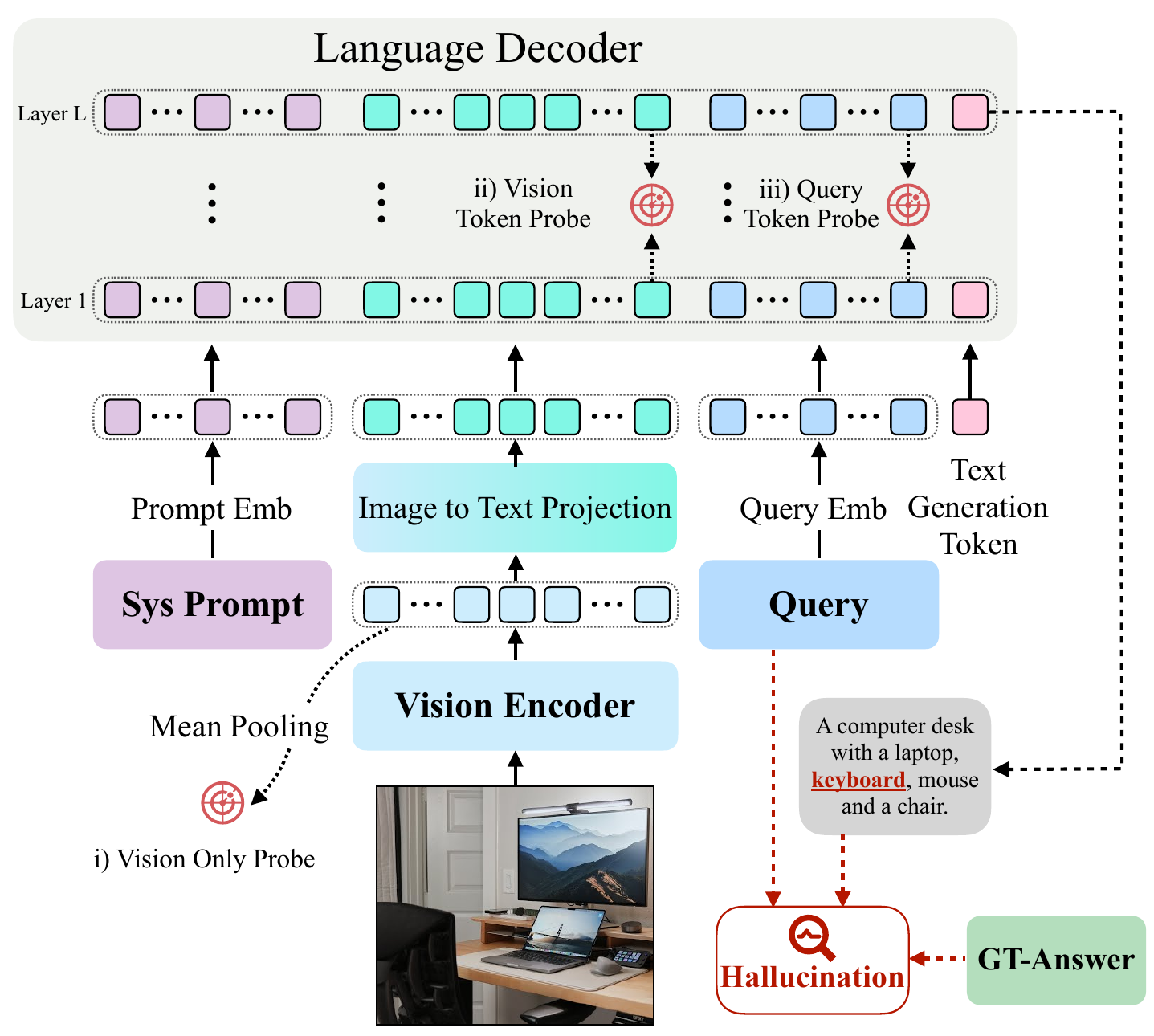}
    \caption{HALP detection pipeline. We extract three types of representations: (1) visual feature vectors from the encoder, (2) vision token states from decoder layers at the final vision patch position, and (3) query token states from the same layers at the final query position. These representations are used independently to probe for hallucination prediction prior to decoding.}
    \label{fig:architecture}
\end{figure}

Next we describe how {\name} explores simple probes on various VLM features for hallucination detection. An overview of the pipeline is shown in Figure~\ref{fig:architecture}.

\subsection{Hallucination Evaluation}
\label{sec: Hullucination Detec}

Hallucination detection in VLMs typically requires generating \emph{complete} responses and comparing them against the ground truth or visual evidence to identify factual inconsistencies.
Prior research has adopted different VLM hallucination evaluation protocols, such as CHAIR \citep{rohrbach-etal-2018-object} and POPE \citep{li2023pope} that rely on the ground truth, and FaithScore \citep{jing2024faithscore} that relies on the visual evidence.

To explore how pre-generation internal features encode information for detecting whether VLMs hallucinate on their visual inputs, we first systematically evaluate model hallucinations through a unified assessment process.

Given a collection of $(I^j, Q^j)$ testing examples along with their ground truth answers or reference generations ${Y^j}$, we run each VLM using the model's standard inference procedure to complete its generation $\hat{Y}^j$ given inputs $(I^j, Q^j)$.
We then employ an automated evaluation framework to determine the hallucination occurrence, denoted as $b^j\in\{0,1\}$ which is a binary label indicating whether any hallucination occurred in the generated response $\hat{Y}^j$.

In particular, we use LLM-based judging \citep{zheng2023judging} to ensure reliable and standardized hallucination detection across diverse queries and visual domains.
Formally,
$\mathrm{LLM}_{\mathrm{judge}}(\hat{Y}^j, Y^j, Q^j) \mapsto b^j$, where we purpose an advanced large language model $\mathrm{LLM}_{\mathrm{judge}}$ to compare the generated response $\hat{Y}^j$ against the ground truth $Y^j$ while considering the query $Q^j$, to identify whether there exists hallucinations such as factual inconsistencies, object/relation/attribute misinterpretations, etc.
The obtained hallucination labels $b^j$ will serve as targets for our probing. Experimental discussions on the reliability of LLM-as-a-judge in our context can be found in Section~\ref{sec:llm-judge-eval}.

\subsection{Pre-Generation Feature Extraction}
\label{subsec:feature}

{\name} extracts three distinct types of internal representations (see Figure~\ref{fig:architecture}) from a \emph{single} forward pass through the VLM before response token generation, each capturing different aspects of visual understanding and multimodal reasoning:

\noindent\textbf{Visual Features (VF)}\,\,
The globally mean pooled output from the vision encoder after processing the image $I$ but before the multimodal projection layer $\mathcal{M}$: $\bar{\mathbf{u}}=\frac{1}{M}\sum_{i=1}^M\mathbf{u_i}$.
This representation captures pure visual information before any LLM and textual integration. Hallucination detection based on this directly probes perception-based signals.

\noindent\textbf{Vision Token Representations (VT)}\,\,
Hidden states from the decoder LLM layer $\ell$ at the \emph{final position} of the visual token sequence $V$. These representations capture how visual information is processed and integrated within the multimodal text decoder. We extract these from five strategically selected layers: %
$\ell \in \{1, \lfloor L/4 \rfloor, \lfloor L/2 \rfloor, \lfloor 3L/4 \rfloor, L\}$, where $L$ is the total number of decoder layers and could be different for different VLMs.

\noindent\textbf{Query Token Representations (QT)}\,
Hidden states from the decoder LLM layer $\ell$ at the \emph{final position} of the query token sequence. Note that this is the final position of the concatenated sequence $(V, Q)$, integrating both vision tokens and text query tokens.
These representations encode the fully contextualized multimodal information that directly precedes text generation.
We extract them from the same five layers as in VT above.

\subsection{Probing}

For each representation type at a particular layer, we train a lightweight probe individually to assess its potential of hallucination detection before generation.
Each probe is a 3-layer MLP with hidden dimensions [512, 256, 128], using ReLU activations. The final layer outputs a binary classification label for the presence of hallucinations $b^j$, indicated by a probe score $s^j\in [0, 1]$, the higher the more likely hallucination will happen.
More details in probe training and hallucination detection evaluation can be found in Section~\ref{sec:experimental-setup}.

\begin{figure*}[!t]
    \centering
    \includegraphics[width=0.99\linewidth]{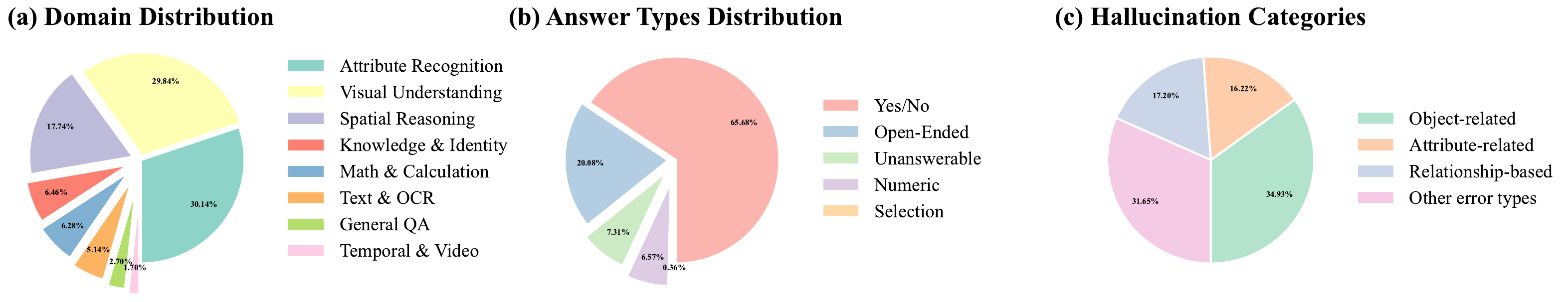}
    \caption{Hallucination detection dataset distributions of task domains, answer formats, and hallucination question types.}
    \vspace{-10pt}
    \label{fig:Dataset}
\end{figure*}

\section{Data and Experiments}

\subsection{Data Construction}

We assemble a dataset of 10,000 samples of image and QA pairs from six established VQA benchmarks to ensure broad coverage of vision-language challenges. Detailed data composition can be found in Table~\ref{tab:eval-datasets}. Figure~\ref{fig:Dataset}(a) shows the domain breakdown: Attribute Recognition and Visual Understanding dominate (60\% combined), with substantial contributions from Spatial Reasoning (17.7\%) and smaller but critical subsets in Knowledge \& Identity, Math \& Calculation, Text \& OCR, General QA, and Temporal \& Video. Figure~\ref{fig:Dataset}(b) presents the answer format distribution: Yes/No questions comprise two‐thirds of the samples, while Open‐Ended, Unanswerable, Numeric, and Selection types provide diverse reasoning and generation tasks. Finally, Figure~\ref{fig:Dataset}(c) categorizes hallucination types tested by the questions:\footnote{Claude Opus 4.0 is used to categorize these questions.} Object-related questions are most frequent (34.9\%), followed by Other questions types (31.7\%), Relationship-based (17.2\%), and Attribute-related (16.2\%) questions. Together, these distributions validate that our dataset captures the variety of tasks and error modes necessary for rigorous hallucination detection evaluation.

\begin{table}[t]
\centering
\small
\resizebox{\columnwidth}{!}{
\begin{tabular}{l l r}
\toprule
\textbf{Dataset} & \textbf{Focus} & \textbf{Samples} \\
\midrule
AMBER \citep{an2023amber}          & discriminative tasks     & 3,926 \\
HaloQuest \citep{cheng2023haloquest} & adversarial challenges   & 2,784 \\
POPE \citep{li2023pope}            & object hallucination     & 1,230 \\
MME \citep{fu2023mme}              & multimodal reasoning     &   885 \\
HallusionBench \citep{guan2023hallusionbench} & visual illusions &   617 \\
MathVista \citep{lu2023mathvista}  & mathematical reasoning   &   558 \\
\midrule
\textbf{Total} & & \textbf{10,000} \\
\bottomrule
\end{tabular}
}
\caption{Our hallucination detection dataset composition.}
\label{tab:eval-datasets}
\end{table}

\subsection{Experimental Setup}
\label{sec:experimental-setup}

\paragraph{Models}
We quantitatively evaluate {\name}'s effectiveness across eight contemporary VLMs spanning diverse architectures and parameter scales: 
Gemma3-12B%
\footnote{\url{huggingface.co/google/gemma-3-12b-it}}
\citep{steiner2024paligemma2familyversatile}, 
LLaVA-1.5-8B%
\footnote{\url{huggingface.co/llava-hf/llava-1.5-7b-hf}}
\citep{liu2024llavanext}, 
Llama-3.2-11B%
\footnote{\url{huggingface.co/meta-llama/llama-3.2-11B-Vision-Instruct}}
\citep{meta2024llama32vision}, 
Phi4-VL-5.6B%
\footnote{\url{huggingface.co/microsoft/Phi-4-multimodal-instruct}}
\citep{li2022blip}, 
Molmo-7B%
\footnote{\url{huggingface.co/allenai/Molmo-7B-0-0924}}
\citep{molmo2023}, 
Qwen2.5-VL-7B%
\footnote{\url{huggingface.co/Qwen/Qwen2.5-VL-7B-Instruct}} \citep{bai2024qwen25vl}, 
SmolVLM2-2.2B%
\footnote{\url{huggingface.co/HuggingFaceTB/SmolVLM2-2.2B-Instruct}}
\citep{smolvlm2024} and 
FastVLM-7B%
\footnote{\url{huggingface.co/apple/FastVLM-7B}} 
\citep{fastvlm2024}.

\paragraph{Post-Generation Hallucination Detection}
We employ an LLM-as-a-judge framework for automated hallucination assessment follows section~\ref{sec: Hullucination Detec}. This automated evaluation maintains consistency across our 10,000-sample dataset while enabling scalable hallucination assessment.

\paragraph{Hallucination Detection Evaluation} We use AUROC (Area Under Receiver Operating Characteristic curve) as our primary metric, which provides a threshold-independent measure of binary classification performance. AUROC ranges from 0 to 1; 0.5 corresponds to random chance, 1 indicates perfect prediction, and values below 0.5 indicate an inverted ranking, making it ideal for comparing hallucination prediction across different models and representation types.

\paragraph{Training Configuration} Probe training was performed with the Adam optimizer (learning rate = 0.001), a batch size of 32, and 50 epochs. The dataset was split 80/20 for training and validation using stratified sampling to preserve the same proportion of different hallucination type examples in each set, ensuring that the validation fold reflects the overall hallucination type.
All experiments use random seed 42 for reproducibility.
All experiments are conducted on NVIDIA RTX 4090 GPUs with 24GB VRAM. We use PyTorch 2.0 with CUDA 11.8 for efficient computation. Model representations are extracted using precision fp16 to optimize memory usage while maintaining numerical stability. 
End-to-end processing of the 10k benchmark data across 8 VLMs (prefill + hidden-state extraction) and training all probes took approx. 10 GPU-hours on a single RTX 4090.

\section{Results and Analysis}

\begin{table}[t]
\centering
\small
\resizebox{\columnwidth}{!}{%
\begin{tabular}{lccc|c}
\toprule
 \textbf{Model Name}  & \textbf{VF} & \textbf{VT} & \textbf{QT} & \textbf{Average}\\

\midrule
Gemma3-12B & 0.6736 & 0.5956 & \textbf{0.9349} & 0.7347\\
FastVLM-7B & 0.6830 & 0.7028 & 0.6136 & 0.6665\\
LLaVa-Next-8B & 0.6108 & 0.6270 & 0.9026 & 0.7135\\
Molmo-V1-7B & 0.6830 & 0.6867 & 0.9193 & 0.7630\\
Qwen2.5-VL-7B & \textbf{0.7873} & 0.6683 & 0.9150 & 0.7902\\
Llama-3.2-11B-Vision & 0.7703 & 0.7377 & 0.8959 & 0.8013\\
Phi4-VL-5.6B & 0.6166 & \textbf{0.7738} & 0.9033 & 0.7646\\
SmolVLM2-2.2B & 0.7238 & 0.6894 & 0.9014 & 0.7715\\
\midrule
\textbf{Average} & 0.6935 & 0.6852 & 0.8733 & 0.7507 \\
\bottomrule
\end{tabular}%
}
\caption{AUROC performance for hallucination detection probes across different vision-language models and different model representations.
}
\label{tab:vlm_comparison}
\end{table}

\subsection{Main Results}
Table~2 summarizes the key results, with VT and QT features extracted from the final decoder layer.\footnote{As our default setup for VT \& QT when layer unspecified.} Across the evaluated models, three consistent patterns emerge.

First, \textit{query token (QT) representations dominate across models}, with seven out of eight models achieving peak AUROC between 0.90 and 0.94 when classification is based on deep query states.\footnote{Rounding off Llama-3.2-11B-Vision's 0.8959 to 0.90.}

Second, \textit{architectural heterogeneity is evident in optimal information encoding}. In some models, visual features already yield strong hallucination detection (Qwen2.5-VL: 0.7873 AUROC; Llama-3.2-11B: 0.7703), suggesting balanced encoding from perception onward. In contrast, models such as LLaVA-Next and Phi4-VL perform poorly with visual features features (0.6108, 0.6166 AUROC), depending more on multimodal integration for robust prediction. \textit{FastVLM-7B illustrates architectural peculiarity}, with maximum AUROC (0.7093) obtained from vision token features rather than from the query position, suggesting a preference for fusion at earlier reasoning stages.

Third, when comparing all representation types, query token vectors consistently provide the most reliable hallucination prediction, with an average AUROC of 0.8898. This outperforms vision tokens (0.7090) and visual features (0.6955), highlighting that the ideal intervention point for hallucination detection is just after multimodal reasoning but prior to text generation. Vision token features themselves show inconsistent value, plateauing at around 0.70 AUROC, with only Llama-3.2-11B and Phi4-VL exceeding 0.75. Pure vision representations act as a competitive baseline, especially in vision-grounded models, but there remains a notable gap between visual features and query-based performance ($\Delta$AUROC $\approx$ 0.15--0.25), demonstrating the importance of multimodal reasoning components.

Overall, our results establish {\name} as a robust framework for detecting hallucination-predictive signals across a range of VLM architectures. Query token probes consistently provide the most effective means for pre-generation assessment, supporting the integration of such methods in safety-critical applications that require rigorous risk evaluation prior to text output.

\subsection{Layer-wise Performance of Hallucination Detection}

\begin{figure*}[htbp]
    \centering
    \includegraphics[width=0.95\textwidth]{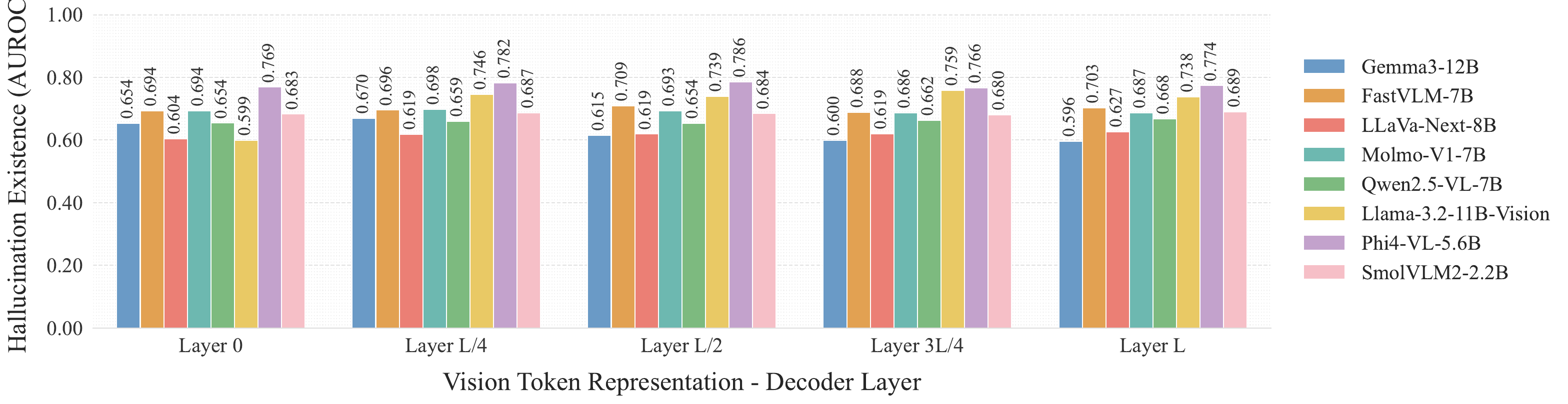}
    \caption{AUROC scores for the final Vision Token (VT) representation at different decoder layers.}
    \vspace{-10pt}
    \label{fig:vision-token-auroc}
\end{figure*}

\begin{figure*}[htbp]
    \centering
    \includegraphics[width=0.95\textwidth]{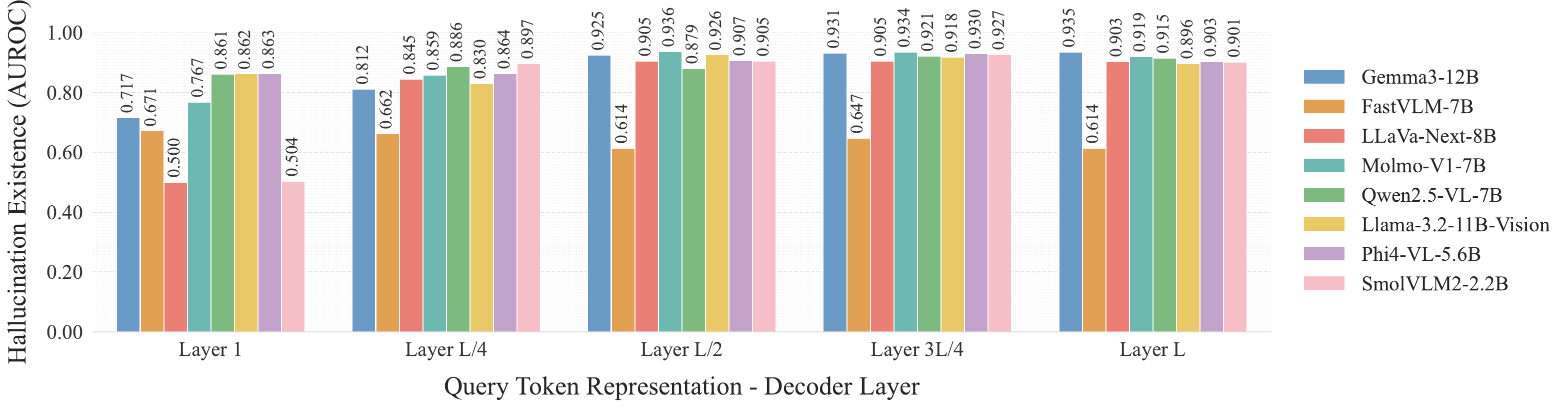}
    \caption{AUROC scores for the final Query Token (QT) representation at different decoder layers.}
    \vspace{-10pt}
    \label{fig:query-token-auroc}
\end{figure*}

Our layer-wise analysis, demonstrated in  Figures \ref{fig:vision-token-auroc} and \ref{fig:query-token-auroc}, reveals distinct patterns in how hallucination-predictive signals evolve across decoder depths, $\ell \in \{1, \lfloor L/4 \rfloor, \lfloor L/2 \rfloor, \lfloor 3L/4 \rfloor, L\}$, where $L$ is the decoder depth.
Significant architectural variations in optimal extraction layers are observed.
Extraction depth for optimal query token performance varies: certain models, such as Gemma3-12B, deliver maximum AUROC at the final layer (0.9349 at $L$), while others like Molmo-V1 are best at intermediate decoding depths (0.9365 at $L/2$).
The analysis demonstrates two fundamentally different progression patterns depending on the representation type and model architecture. Query token representations generally show substantial improvement from shallow to deep layers, while vision token representations exhibit more stable but limited performance across all depths.

\paragraph{Vision Token Layer-wise Behavior}

As shown in Figure~\ref{fig:vision-token-auroc}, vision token (VT) representations show markedly different patterns, with generally stable performance across layers and limited overall improvement. Most models plateau around 0.65-0.70 AUROC across all depths, suggesting that visual information processed within decoder layers provides consistent but limited hallucination prediction capability. Notably, Phi4-VL achieves the strongest vision token performance, maintaining 0.77-0.78 AUROC across layers, while FastVLM-7B peaks at Layer $L/2$ (0.7093), aligning with its unique architectural behavior identified in our cross-model analysis.

\paragraph{Query Token Layer-wise Progression}

Shown in Figure~\ref{fig:query-token-auroc}, query token (QT) representations demonstrate remarkable layer-wise improvement across most models, with AUROC scores increasing substantially as layer increases. Most models achieve their peak performance in deep layers ($L/2$ to $L$), with dramatic improvements from early layers. For instance, 
Gemma-3-12B exemplifies the typical progression pattern, with consistent improvement across all layers: $0.7165 \to 0.8119 \to 0.9247 \to 0.9315 \to 0.9349$, showing how hallucination signals become increasingly concentrated as the model approaches final output generation. This monotonic improvement suggests that multimodal reasoning gradually refines hallucination-relevant features through successive Transformer layers.

\vspace{-5pt}

\paragraph{Layer-wise Optimization Insights}

The optimal extraction layers vary significantly across models and representation types. For query tokens, layer $3L/4$ emerges as a consistently strong choice, achieving peak or near-peak performance for most models (Gemma-3: 0.9315, Phi4-VL: 0.9305, SmolVLM: 0.9272). This suggests that the penultimate reasoning stage, just before final output commitment, contains maximal hallucination-predictive information.

Vision tokens show more model-specific optimal layers: FastVLM-7B peaks at layer $L/2$ (0.7093), Phi4-VL at layer $L/2$ (0.7858), and Llama-3.2-11B-Vision at layer $3L/4$ (0.7592). This heterogeneity indicates that different architectures process and preserve visual hallucination signals at varying depths, likely reflecting diverse approaches to visual-textual integration.

These layer-wise patterns provide crucial insights for practical deployment: practitioners can optimize computational efficiency by extracting representations at optimal layers rather than defaulting to final outputs, while the consistent strength of mid-to-deep query token representations offers robust hallucination prediction across diverse architectural choices.

\begin{table*}[h]
\centering
\resizebox{\textwidth}{!}{%
\begin{tabular}{l c c c c c c c c c}
\toprule
\textbf{Hallucination Type} & \textbf{Count} & \textbf{FastVLM-7B} & \textbf{Gemma3-12B} & \textbf{LLaVA-Next-8B} & \textbf{Llama-3.2-11B} & \textbf{Molmo-V1} & \textbf{Phi4-VL} & \textbf{Qwen2.5-VL-7B} & \textbf{SmolVLM2-2.2B} \\
\midrule
Attribute-Related & 3493 & 7.58\%  & 11.84\% & 19.73\% & 9.58\%  & 14.30\% & 8.51\%  & \textbf{4.87\%}  & 20.47\% \\
Object-Related  & 1622  & \textbf{5.81\%}  & 10.42\% & 8.93\%  & 7.27\%  & 10.91\% & 6.80\%  & 6.38\%  & 8.33\%  \\
Relationship    & 1720  & 6.74\%  & 5.12\%  & 15.29\% & 4.91\%  & 7.44\%  & 3.73\%  & \textbf{0.87\%}  & 2.62\%  \\
Other           &  3165 & 13.52\% & 13.08\% & 14.66\% & 14.46\% & 14.66\% & 16.85\% & \textbf{9.13\%}  & 10.21\% \\
\bottomrule
\end{tabular}%
}
\caption{Hallucination rates by basic hallucination type across eight vision–language models. Bold values indicate the lowest hallucination rate among tested models per category.}
\label{tab:hallucination_rates_horizontal}
\end{table*}

\subsection{Hallucination Types and Application Domains}

We analyze the hallucination prediction with {\name} on different domains and types of the evaluation questions, categorized in Figure~\ref{fig:Dataset}(a) and (c).

\paragraph{Hallucination Types}
Table~\ref{tab:hallucination_rates_horizontal} shows the hallucination rates among different VLMs when tested on different types of visual understanding.
Hallucination prediction performance for different error types is shown in Table~\ref{tab:hallucination_types}.
Attribute-Related and Other errors are the most frequent and hardest to detect, while Object-Related errors though less common yield strong QT signals (AUROC 0.878). Relationship errors are rare (6.06 \%) but notably difficult to predict (AUROC 0.708).
Across all categories, query-token (QT) probes consistently outperform vision-only (VF) representations, with the largest AUROC gaps for Attribute (+0.237) and Relationship (+0.149) errors. 
These examples demonstrate how query-token (QT) representations effectively identify hallucination risks across diverse error types before text generation.

\begin{figure}[t]
    \centering
    \includegraphics[width=1.0\linewidth]{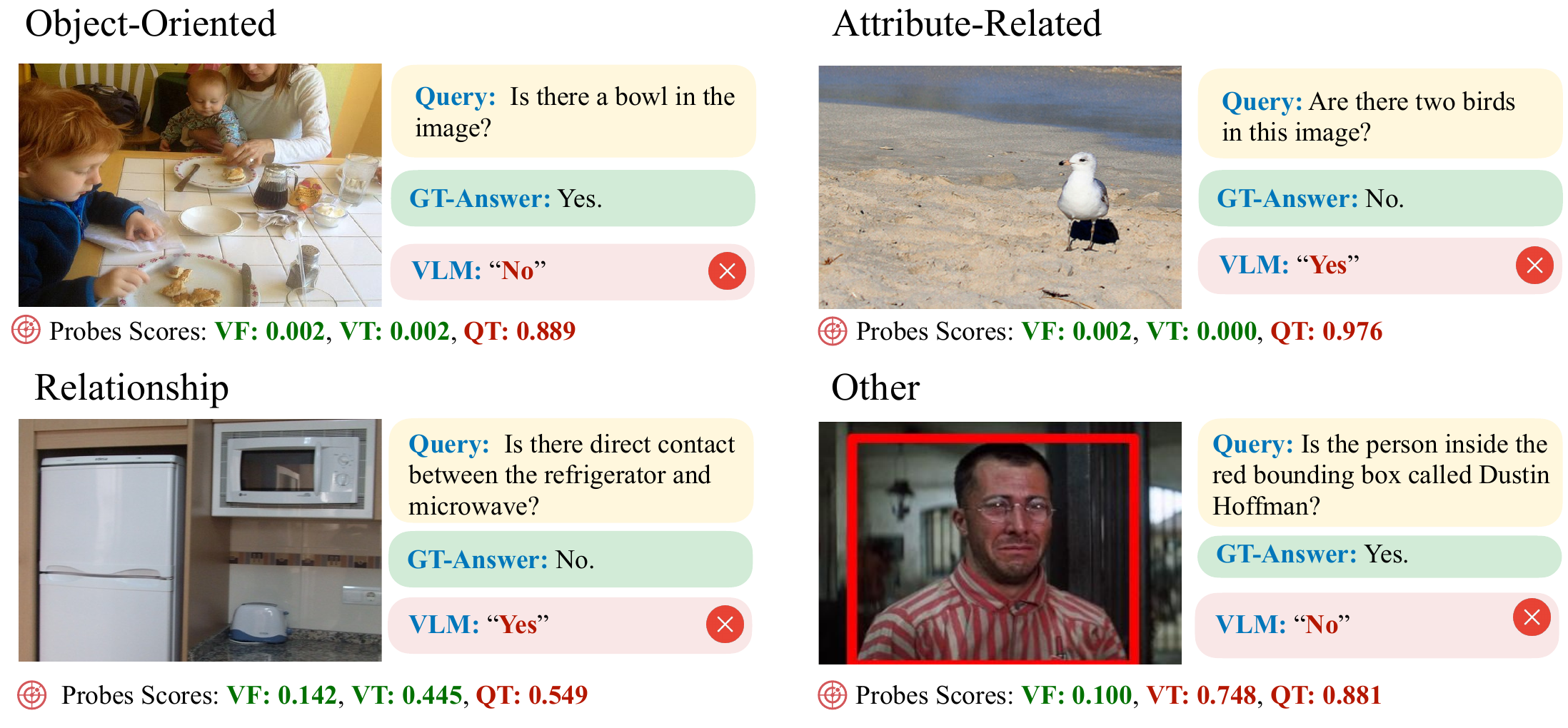}
    \caption{Hallucination types and probe scores to predict possible hallucinations before decoding, with Qwen2.5-VL-7B. Higher scores indicate greater chance of answer hallucination.}
    \label{fig:quantitative}
\end{figure}

Figure~\ref{fig:quantitative} further illustrates examples of the four hallucination categories—Object-Oriented, Attribute-Related, Relationship, and Other—along with image (I), question (Q), ground truths (\(Y^j\)), model answers (\(\hat{Y}^j\)),  and probe scores ($s^j$) for vision-only (VF), vision-token (VT), and query-token (QT) representations.

\paragraph{Application Domains}
Table~\ref{tab:domain_performance} shows the average hallucination rates and {\name} prediction performance with QT representations before generation among all models.
Some concrete examples can be seen in Figure~\ref{fig:domain-type}.
High-risk domains such as Temporal \& Video (46.9 \% error rate, QT AUROC 0.456) and Knowledge \& Identity (21.2 \%, 0.694) are the weakest areas that require enhanced monitoring, whereas low-risk domains Attribute Recognition (7.1 \%, 0.863), Text \& OCR (6.9 \%, 0.855), and Visual Understanding (10.3 \%, 0.859) exhibit robust performance, suitable for automated deployment.

\begin{table}[t]
\centering
\resizebox{\columnwidth}{!}{%
\begin{tabular}{lcccc}
\toprule
\textbf{Hallucination Type} & \textbf{Rate (\%)} & \textbf{VT AUROC} & \textbf{QT AUROC} & \textbf{$\Delta$AUROC} \\
\midrule
Attribute-Related & 11.92 & 0.567 & 0.804 & 0.237 \\
Object-Related    & 7.94  & 0.734 & 0.878 & 0.144 \\
Relationship      & 6.06  & 0.559 & 0.708 & 0.149 \\
Other             & 13.47 & 0.755 & 0.827 & 0.072 \\
\bottomrule
\end{tabular}%
}
\caption{Hallucination rates and AUROC performance by error type averaged across models. $\Delta$AUROC shows the performance gain from query token (QT) over visual features (VF) representations.}
\label{tab:hallucination_types}
\end{table}

\begin{table}[t]
\centering
\small
\resizebox{\columnwidth}{!}{%
\begin{tabular}{lcc}
\toprule
\textbf{Domain} & \textbf{Rate (\%)} & \textbf{QT AUROC} \\
\midrule
Temporal \& Video      & 46.90 & 0.456 \\
Knowledge \& Identity  & 21.19 & 0.694 \\
Math \& Calculation    & 16.79 & 0.819 \\
Visual Understanding   & 10.26 & 0.859 \\
Attribute Recognition  & 7.14  & 0.863 \\
General QA             & 7.04  & 0.797 \\
Text \& OCR            & 6.94  & 0.855 \\
Spatial Reasoning      & 5.85  & 0.713 \\
\bottomrule
\end{tabular}%
}
\caption{Hallucination rates and query token AUROC performance by application domain, ordered by hallucination rate.}
\label{tab:domain_performance}
\end{table}

\subsection{LLM-as-a-Judge Evaluation}
\label{sec:llm-judge-eval}

We adopt LLM-as-a-judge with GPT-4 \citep{zheng2023judging} to label hallucinations. To verify its reliability, three expert annotators independently labeled a stratified sample of 200 examples, achieving substantial inter-annotator agreement with a Fleiss value of 0.89. The GPT-4 judge matched human consensus on 97\% of cases, confirming its suitability for large-scale, automated hallucination labeling in our multi-domain evaluation.

\section{Discussion}

\subsection{Architectural Heterogeneity}
\label{sec:interpretation}

Our experiments reveal substantial architectural heterogeneity in {\name}: the most predictive representation type and layer are not universal across VLMs. We propose two hypotheses (\textbf{H}) to explain the observed layer- and architecture-wise trends.

\paragraph{H1: Semantic-to-task transition.}
Across several models, query-token representations peak in predictive performance at intermediate layers (e.g., around $L/2$ or $3L/4$; Figure~\ref{fig:query-token-auroc}). We hypothesize that intermediate layers retain richer multimodal semantic structure, while later layers become increasingly specialized toward next-token prediction, which can compress or attenuate auxiliary signals such as internal uncertainty or grounding misalignment. This provides a plausible explanation for cases where mid-layer query tokens outperform the final layer.

\paragraph{H2: Vision-centric vs.\ fusion-centric locus of grounding.}
We observe that some architectures exhibit strong hallucination predictability from visual-only features, while others rely 
predominantly on post-fusion query-token states (Figure~\ref{fig:vision-token-auroc}).
We hypothesize that ``vision-centric'' models encode stronger perceptual grounding signals earlier in the vision stack (making perceptual failures detectable before fusion), whereas ``fusion-centric'' models encode the most informative risk signals after multimodal integration, consistent with failures driven by linguistic--visual alignment rather than raw perception. This suggests that the optimal monitoring point depends on where grounding is implemented in the architecture.

\subsection{From Detection to Control}
\label{sec:control}

While {\name} is primarily diagnostic, its pre-generation risk score can directly serve as a control signal. We outline two lightweight mitigation policies early refusal/deferral and selective routing that can be implemented without modifying the underlying VLM, and discuss how they leverage {\name} scores in practical deployments.

\vspace{-0.1in}
\paragraph{Early refusal / deferral.}
The {\name} score can be used to decide when the system should respond versus when it should decline or defer. Concretely, inputs can be ranked by predicted hallucination risk; the system returns a refusal/deferral response (e.g., ``I am unsure'' or a request for clarification) for the highest-risk cases and answers the remaining lower-risk inputs normally. This provides a simple, threshold-based safeguard that trades response coverage for higher reliability.
A preliminary study is presented in Appendix~\ref{appendix:threshold}.

\vspace{-0.1in}
\paragraph{Selective routing.}
{\name} scores can also be used for hierarchical deployments in which high-risk inputs are routed from a base VLM to a stronger VLM (or a tool-augmented pipeline), while low-risk inputs are answered by the base model. This provides a simple mechanism to concentrate additional compute on uncertain cases while keeping average latency low.

\vspace{-0.1in}
\paragraph{Runtime overhead.}
In deployment, {\name} requires extracting a single chosen representation layer and applying a small probe \textit{during the prefill stage} (single forward pass). We confirm that extracting hidden states from the decoder during a single forward pass does not introduce extra passes. In our NVIDIA RTX 4090 GPU testbed, the probe inference time for a 3-layer MLP was consistently between 10-15ms. For reference, typical end-to-end generation (prefill + decoding 100 tokens) is substantially larger, so the relative overhead is <1\% when compared against full generation.

\subsection{Practical Implications}
\label{sec:Implications}
The analysis reveals distinct risk profiles for different application contexts. \textit{High-risk domains} (Temporal \& Video, Knowledge \& Identity) require enhanced monitoring and potentially specialized architectures, while \textit{moderate-risk domains} (Math \& Calculation, Visual Understanding) can benefit from standard {\name} deployment with appropriate thresholds. \textit{Low-risk domains} (Attribute Recognition, Text \& OCR, Spatial Reasoning) demonstrate reliable performance and can support more automated deployment scenarios. However, even in low-risk domains, the substantial performance improvement from query token over visual features probes (average improvement of 0.2-0.3 AUROC) \textit{justifies the computational overhead of deeper representation extraction}.

These domain-specific patterns provide actionable guidance for practitioners: applications focused on temporal reasoning or factual knowledge retrieval should implement additional safeguards, while visual recognition and OCR tasks can rely more confidently on {\name}'s pre-generation predictions.

\vspace{-5pt}
\section{Conclusion}
\vspace{-5pt}
We presented {\name}, a lightweight framework for pre-generative prediction of hallucinations in vision language models (VLMs). Through extensive experiments across eight modern VLMs and diverse multimodal benchmarks, we examine representations from global visual features to decoder-level vision and query token states within a single forward pass prior to decoding. Our findings show that hallucination risk is detectable before generation and that the most informative signals vary across architectures. The approach enables rapid, real-time hallucination risk assessment without costly decoding. Future work includes integrating probe-based risk scores into decoding-time control for on-the-fly hallucination prevention, adaptive routing, and confidence-aware generation, as well as extending the approach to finer-grained detections.

\clearpage

\section*{Ethical Considerations}

This work addresses hallucination detection in vision-language models, which carries several important ethical implications that must be carefully considered.

\paragraph{Beneficial Applications and Risk Mitigation}
Our hallucination detection framework serves beneficial purposes by enhancing AI safety and reliability. By enabling early detection of model hallucinations, this research contributes to preventing the spread of misinformation and reducing potential harm from AI-generated false content. The probing approach offers a transparent method to assess model reliability without requiring access to proprietary architectures, supporting broader AI safety research.

\paragraph{Limitations and Misuse Potential}
While designed for safety enhancement, this technology could potentially be misused. Adversarial actors might attempt to circumvent hallucination detectors or exploit knowledge of detection mechanisms to craft more sophisticated deceptive content. Additionally, over-reliance on automated hallucination detection without human oversight could lead to false confidence in AI system outputs.

\paragraph{Dataset and Evaluation Concerns}
Our evaluation relies on existing VQA benchmarks, which may contain inherent biases or limitations in representing diverse populations and scenarios. The use of GPT-4 as a judge for hallucination labeling introduces potential bias from the judging model itself, though we employ lenient criteria to minimize false positives. Future work should incorporate more diverse evaluation frameworks and human annotation validation.

\paragraph{Transparency and Reproducibility}
We commit to open research practices by providing detailed methodology and encouraging reproducible research. However, we acknowledge that computational resource requirements may limit accessibility for some researchers. The internal representation analysis approach requires significant computational infrastructure, potentially creating barriers to widespread adoption.

\paragraph{Broader Impact}
This research contributes to the responsible development of AI systems by providing tools to assess and improve their reliability. However, we emphasize that hallucination detection should complement, not replace, human judgment in high-stakes applications. Continued research into AI safety, interpretability, and robustness remains essential for the responsible deployment of vision-language models in real-world applications.

\section*{Limitations}

While our probing framework effectively detects a range of hallucination types, several limitations remain. First, the reliance on VQA benchmarks may introduce dataset biases, limiting generalization to other vision–language tasks and real-world scenarios. Second, GPT-4 based judging, despite lenient criteria, can inherit biases and occasional misclassifications from the judge model, affecting label quality. Third, our approach focuses on pre-generation internal signals and does not directly address post-generation mitigation strategies. Fourth, high computational cost for extracting and probing internal representations may hinder deployment in resource-constrained settings. Fifth, the current taxonomy covers major hallucination types but may not capture nuanced cases, such as context-dependent or culturally specific errors, which warrant further study. Finally, our experiments use small-to-medium VLM sizes; performance characteristics and hallucination patterns in larger models remain underexplored.

\clearpage

\bibliography{reference}

\clearpage %

\appendix
\raggedbottom

\section*{Appendix}
\label{sec:appendix}

\section{Model Analysis}

This appendix provides a detailed breakdown of the performance metrics for each vision-language model (VLM) evaluated in this study. For each model, we present the AUROC score for the binary classification task. The results are organized by the type of internal representation used for the probe: the Visual Features (VF), last Vision Token Representation (VT) and last Query Token Representation (QT) at various decoder layers.

\begin{figure}[t]
    \centering
    \includegraphics[width=0.9\linewidth]{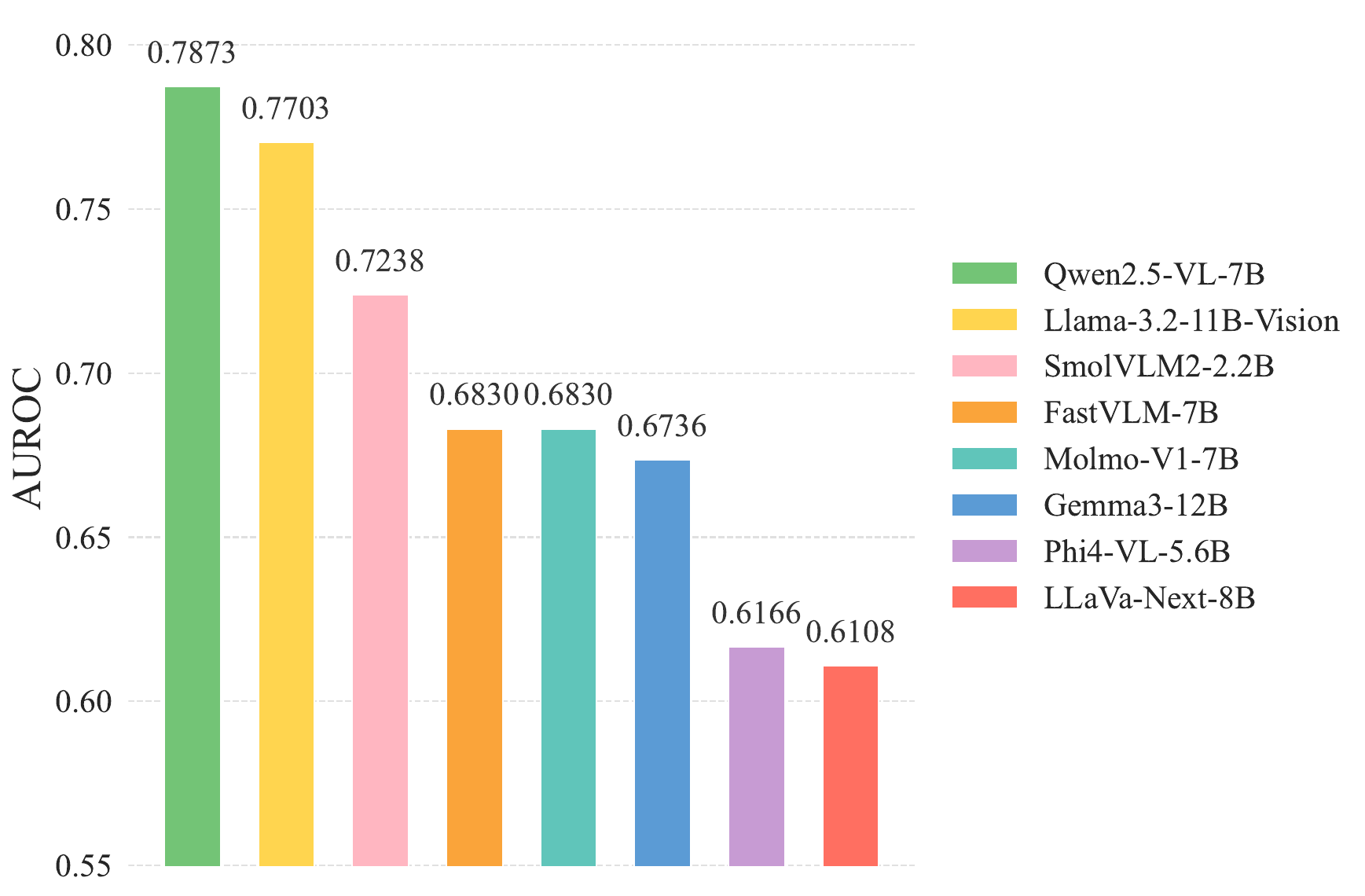}
    \caption{AUROC scores for visual features detection.}
    \label{fig:visual features-auroc}
\end{figure}

Besides Table~\ref{tab:vlm_comparison}, Figure~\ref{fig:vision-token-auroc} and Figure~\ref{fig:query-token-auroc}, detailed results are supplemented in Table~\ref{tab:combined_auroc_scores}, Table~\ref{tab:combined_auroc_scores_2}, and Figure~\ref{fig:visual features-auroc} (hallucination prediction performance with VF representation only before the decoder, sorted by the models).

\paragraph{Gemma3-12B}
The results for Gemma3-12B demonstrate an apparent shift in hallucination predictive power from vision-based to query token representations across decoder layers. As shown in Table~\ref{tab:combined_auroc_scores}, query token probes significantly outperform both visual features and vision token approaches.

\begin{itemize}
\item \textbf{Query Token Dominance}: Query token representations show exceptional performance, with AUROC increasing progressively through deeper layers from 0.7165 (Layer $1$) to 0.9349 (Layer $L$). This suggests that hallucination signals become increasingly concentrated in the final reasoning states before generation.

\item \textbf{Vision Representations Plateau}: Both visual features (0.6736) and vision token representations (peak: 0.6698 at Layer $L/4$) show moderate but limited predictive power. Vision token performance actually degrades in deeper layers, declining from 0.6698 to 0.5956.

\item \textbf{Layer-wise Progression}: The most striking finding is the consistent improvement in query token performance with depth, achieving near-perfect detection (0.9349 AUROC) at the final layer, indicating that Gemma3-12B's hallucination signals are most accessible just before text generation begins.
\end{itemize}

\paragraph{FastVLM-7B}
FastVLM-7B exhibits a unique architectural behavior that diverges significantly from other models in the study. As shown in Table~\ref{tab:combined_auroc_scores}, this model demonstrates an unusual pattern where vision token representations outperform both visual features and query token approaches.

\begin{itemize}
\item \textbf{Vision Token Superiority}: Vision token representations achieve the highest performance, peaking at 0.7093 AUROC in Layer $L/2$. This suggests that FastVLM-7B's hallucination signals are most accessible when visual information is processed within the decoder context at intermediate depths.

\item \textbf{Query Token Limitations}: Unlike other models, query token representations show consistently poor performance across all layers (0.6136-0.6715), with performance actually degrading in deeper layers. This indicates that hallucination signals do not concentrate in the final reasoning states for this architecture.

\item \textbf{Architectural Anomaly}: The consistent underperformance of query tokens compared to vision tokens (0.7093 vs 0.6715 best scores) suggests that FastVLM-7B processes multimodal information differently, potentially with weaker integration between visual and textual reasoning pathways. This architectural divergence makes it an outlier among contemporary VLMs.
\end{itemize}

\begin{table}[t]
\centering
\resizebox{\columnwidth}{!}{%
\begin{tabular}{llcccccc}
\toprule
\multirow{2}{*}{\textbf{Model Name}} & 
\textbf{Token} & 
\textbf{Vision} &
\multicolumn{5}{c}{\textbf{Decoder Layers}} \\
\cmidrule(lr){4-8}
& \textbf{Type} & \textbf{Encoder} & 
\textbf{Layer 0} & \textbf{Layer $L/4$} & \textbf{Layer $L/2$} & \textbf{Layer $3L/4$} & \textbf{Layer $L$} \\
\midrule
Gemma3-12B   & VF & 0.6736 &   -   &    -    &    -    &    -    &   -     \\
             & VT &    -    & 0.6538 & 0.6698 & 0.6146 & 0.5995 & 0.5956 \\
             & QT &    -    & 0.7165 & 0.8119 & 0.9247 & 0.9315 & 0.9349 \\
\midrule
FastVLM-7B   & VF & 0.6830 &  -      &    -    &     -   &    -    &  -      \\
             & VT &    -    & 0.6935 & 0.6957 & 0.7093 & 0.6879 & 0.7028 \\
             & QT &    -    & 0.6715 & 0.6623 & 0.6139 & 0.6475 & 0.6136 \\
\midrule
LLaVA-Next-8B & VF & 0.6108 &   -    &   -     &   -     &     -   &   -     \\
              & VT &    -    & 0.6037 & 0.6185 & 0.6192 & 0.6194 & 0.6270 \\
              & QT &   -     & 0.4996 & 0.8453 & 0.9049 & 0.9053 & 0.9026 \\
\bottomrule

  SmolVLM-2.2B & VF & 0.7238 & - & - & - & - & - \\                                                                     
               & VT & - & 0.6829 & 0.6868 & 0.6845 & 0.6801 & 0.6894 \\                                                 
               & QT & - & 0.5040 & 0.8971 & 0.9055 & 0.9272 & 0.9014 \\
\bottomrule

\end{tabular}%
}
\caption{Hallucination Detection (AUROC) for probes built on visual features (VF), vision token (VT), and query token (QT) representations at different decoder layers using Gemma3-12B, FastVLM-7B, and LLaVA-Next-8B.}
\label{tab:combined_auroc_scores}
\end{table}

\paragraph{LLaVA-Next-8B}
For LLaVA-Next-8B, query token representations again dominate hallucination detection, with AUROC peaking at 0.9053 in Layer $3L/4$ (Table~\ref{tab:combined_auroc_scores}). visual features and vision token representations offer limited predictive power.

\begin{itemize}
\item \textbf{Query Token Peak}: The best query token performance (0.9053 AUROC) occurs at Layer $3L/4$, closely followed by Layer $L/2$ (0.9049). This indicates that hallucination signals for LLaVA-Next-8B are most accessible just before the final decoding stage.

\item \textbf{Vision Token Marginality}: Vision token representations reach a maximum of 0.6270 AUROC at the final layer, showing marginal improvement over the visual features baseline (0.6108).

\item \textbf{Consistency Across Depths}: Query token performance rises sharply from shallow layers (0.4996 at Layer 1) to deeper layers, highlighting the progressive concentration of hallucination-relevant features in the decoder.
\end{itemize}

 \paragraph{SmolVLM-2.2B}                                                                                              
  SmolVLM-2.2B, the smallest model in our study at 2.2 billion parameters, demonstrates that hallucination detection    
  capability is not strictly dependent on model scale. As shown in Table~\ref{tab:combined_auroc_scores}, query token   
  representations achieve strong performance despite the model's compact size, while exhibiting unique characteristics  
  in vision representation behavior.                                                                                     
  \begin{itemize}                                                                                                       
  \item \textbf{Query Token Peak at Layer $3L/4$}: Unlike larger models where query token performance peaks at the final
   layer, SmolVLM-2.2B achieves its best AUROC of 0.9272 at Layer $3L/4$, with performance slightly declining at Layer  
  $L$ (0.9014). This suggests that hallucination signals are most concentrated slightly before the final decoding stage 
  in this compact architecture.                                               
  \item \textbf{Strong Visual Features}: Notably, visual features (0.7238 AUROC) outperform vision token representations
   across all decoder layers (peak: 0.6894 at Layer $L$). This indicates that SmolVLM-2.2B's vision encoder captures    
  hallucination-predictive signals that are partially diluted during multimodal fusion in the decoder.                  
             
  \item \textbf{Flat Vision Token Performance}: Vision token representations remain relatively stable across all layers 
  (0.6801--0.6894), showing minimal layer-wise variation. This contrasts with larger models that exhibit more pronounced
   layer-dependent patterns in vision token performance.                                                                
       
  \item \textbf{Early Layer Query Limitations}: Query token representations at Layer 0 show near-random performance     
  (0.5040 AUROC), but rapidly improve by Layer $L/4$ (0.8971), demonstrating that hallucination signals emerge quickly  
  in the decoder despite the model's smaller capacity.                                                                  
  \end{itemize}

\paragraph{Molmo-V1-7B}
Molmo-V1-7B exhibits very strong query token performance, with the highest AUROC of 0.9365 at Layer $L/2$ (Table~\ref{tab:combined_auroc_scores_2}).

\begin{itemize}
\item \textbf{Peak Query Token}: Query token probe peaks at Layer $L/2$ with AUROC 0.9365, indicating that hallucination signals are most salient in mid-to-late decoder states.

\item \textbf{Vision Token Moderate}: Vision token representations peak at Layer $L/4$ (0.6982 AUROC) but remain consistently below the query token probe, suggesting moderate predictive power.

\item \textbf{Stable Visual Features}: The visual features probe (0.6830 AUROC) provides a reasonable baseline but is outperformed by all deeper multimodal representations.
\end{itemize}

\paragraph{Qwen2.5-VL-7B}
Qwen2.5-VL-7B achieves its best hallucination detection performance with both visual features and query token probes. As shown in Table~\ref{tab:combined_auroc_scores_2}:

\begin{itemize}
\item \textbf{Strong Visual Features Baseline}: The visual features probe reaches 0.7873 AUROC, indicating robust visual grounding.
\item \textbf{Peak Query Token}: Query token representations peak at Layer $3L/4$ with 0.9215 AUROC, demonstrating the importance of mid-to-deep decoder states.
\item \textbf{Moderate Vision Token}: Vision token representations obtain a maximum of 0.6683 AUROC at the final layer, showing moderate predictive power relative to other representation types.
\end{itemize}

\begin{table}[t]
\centering
\resizebox{\columnwidth}{!}{%
\begin{tabular}{llcccccc}
\toprule
\multirow{2}{*}{\textbf{Model Name}} & 
\textbf{Token} & 
\textbf{Vision} &
\multicolumn{5}{c}{\textbf{Decoder Layers}} \\
\cmidrule(lr){4-8}
& \textbf{Type} & \textbf{Encoder} & 
\textbf{Layer 0} & \textbf{Layer $L/4$} & \textbf{Layer $L/2$} & \textbf{Layer $3L/4$} & \textbf{Layer $L$} \\
\midrule
Molmo-V1-7B   & VF & 0.6830 & -      & -      & -      & -      & -      \\
              & VT & -      & 0.6936 & 0.6982 & 0.6931 & 0.6862 & 0.6867 \\
              & QT & -      & 0.7675 & 0.8588 & 0.9365 & 0.9345 & 0.9193 \\
\midrule
Qwen2.5-VL-7B & VF & 0.7873 & -      & -      & -      & -      & -      \\
              & VT & -      & 0.6543 & 0.6593 & 0.6539 & 0.6625 & 0.6683 \\
              & QT & -      & 0.8614 & 0.8863 & 0.8794 & 0.9215 & 0.9150 \\
\midrule
Llama3.2-11B-Vision & VF & 0.7703 & -      & -      & -      & -      & -      \\
                    & VT & -      & 0.5991 & 0.7463 & 0.7392 & 0.7592 & 0.7377 \\
                    & QT & -      & 0.8623 & 0.8301 & 0.9259 & 0.9178 & 0.8959 \\
\midrule
Phi4-VL       & VF & 0.6166 & -      & -      & -      & -      & -      \\
              & VT & -      & 0.7689 & 0.7821 & 0.7858 & 0.7662 & 0.7738 \\
              & QT & -      & 0.8629 & 0.8638 & 0.9072 & 0.9305 & 0.9033 \\
\bottomrule
\end{tabular}%
}
\caption{Hallucination Detection (AUROC) for probes built on visual features (VF), vision token (VT), and query token (QT) representations at different decoder layers using Molmo-V1-7B, Qwen2.5-VL-7B, Llama3.2-11B-Vision, and Phi4-VL.}
\label{tab:combined_auroc_scores_2}
\end{table}

\paragraph{Llama-3.2-11B-Vision}
Llama-3.2-11B-Vision demonstrates strong performance across multiple representation types. As shown in Table~\ref{tab:combined_auroc_scores_2}, this model achieves excellent visual features performance while maintaining superior query token detection.

\begin{itemize}
\item \textbf{Balanced Performance}: visual features probe achieves 0.7703 AUROC, indicating robust visual grounding. Vision token representations peak at Layer $3L/4$ (0.7592), showing competitive intermediate performance.

\item \textbf{Query Token Peak}: Query token representations achieve maximum performance at Layer $L/2$ (0.9259 AUROC), demonstrating strong hallucination signal concentration in mid-deep decoder states.

\item \textbf{Vision-Grounded Architecture}: The strong visual features baseline suggests that Llama-3.2-11B-Vision encodes substantial hallucination-relevant information at the visual processing stage, making it less dependent on multimodal reasoning for detection.
\end{itemize}

\paragraph{Phi4-VL}
Phi4-VL exhibits unique behavior with strong vision token performance alongside excellent query token detection. Table~\ref{tab:combined_auroc_scores_2} shows distinctive architectural characteristics.

\begin{itemize}
\item \textbf{Vision Token Strength}: Vision token representations perform exceptionally well, peaking at Layer L/2 (0.7858 AUROC), suggesting effective visual-contextual integration within the decoder.

\item \textbf{Query Token Excellence}: Query token representations achieve maximum performance at Layer 3L/4 (0.9305 AUROC), demonstrating strong hallucination prediction capability.

\item \textbf{Visual Features Limitation}: The relatively weak visual features performance (0.6166 AUROC) indicates that Phi4-VL relies heavily on multimodal integration for hallucination detection, rather than pure visual features.
\end{itemize}

\section{Dataset Composition and Characteristics}

Our comprehensive evaluation dataset consists of 10,000 carefully selected image-question pairs that provide a balanced representation of multimodal reasoning challenges and hallucination patterns. The following tables present the detailed breakdown across three key distributional characteristics.

\paragraph{Domain Type Distribution}

\begin{table}[H]
\centering
\resizebox{\columnwidth}{!}{%
\begin{tabular}{l l r}
\toprule
\textbf{Main Domain} & \textbf{Task Categories} & \textbf{Count} \\
\midrule
\multirow{3}{*}{\textbf{Attribute Recognition}} & $\rightarrow$ States (wet/dry, clean/messy) & \multirow{3}{*}{3,014} \\
 & $\rightarrow$ Colors, numbers, counts & \\
 & $\rightarrow$ Physical properties, actions & \\
\midrule
\multirow{3}{*}{\textbf{Visual Understanding}} & $\rightarrow$ Object recognition & \multirow{3}{*}{2,984} \\
 & $\rightarrow$ Existence questions & \\
 & $\rightarrow$ Visual challenges, illusions & \\
\midrule
\multirow{3}{*}{\textbf{Spatial Reasoning}} & $\rightarrow$ Object relationships & \multirow{3}{*}{1,774} \\
 & $\rightarrow$ Positioning, proximity & \\
 & $\rightarrow$ Spatial configurations & \\
\midrule
\multirow{2}{*}{\textbf{Knowledge \& Identity}} & $\rightarrow$ Celebrities, landmarks & \multirow{2}{*}{646} \\
 & $\rightarrow$ Scenes, artwork, common knowledge & \\
\midrule
\multirow{2}{*}{\textbf{Math \& Calculation}} & $\rightarrow$ Numerical reasoning & \multirow{2}{*}{628} \\
 & $\rightarrow$ Table interpretation & \\
\midrule
\multirow{2}{*}{\textbf{Text \& OCR}} & $\rightarrow$ Text reading, OCR & \multirow{2}{*}{514} \\
 & $\rightarrow$ Translation, code understanding & \\
\midrule
\textbf{General QA} & $\rightarrow$ General VQA & 270 \\
\midrule
\multirow{2}{*}{\textbf{Temporal \& Video}} & $\rightarrow$ Video sequences & \multirow{2}{*}{170} \\
 & $\rightarrow$ Temporal ordering & \\
\bottomrule
\textbf{Total} & & \textbf{10,000} \\
\bottomrule
\end{tabular}%
}
\caption{Domain type distribution showing main categories and their specific task subcategories.}
\label{tab:domain_distribution}
\end{table}

The domain distribution as seen in Table~\ref{tab:domain_distribution} reveals a balanced focus on core visual-language capabilities, with \textbf{Attribute Recognition} and \textbf{Visual Understanding} together comprising nearly 60\% of the dataset. This emphasis on fundamental visual perception tasks ensures robust evaluation of VLMs' basic competencies. The inclusion of specialized domains like \textbf{Temporal \& Video} (170 samples) and \textbf{Math \& Calculation} (628 samples) provides targeted assessment of advanced reasoning capabilities, while the substantial representation of \textbf{Spatial Reasoning} (1,774 samples) evaluates models' understanding of object relationships and positioning a critical capability for hallucination detection.

\paragraph{Hallucination Type Distribution}

\begin{table}[H]
\centering
\resizebox{\columnwidth}{!}{%
\begin{tabular}{l l r}
\toprule
\textbf{Main Category} & \textbf{Specific Error Types} & \textbf{Count} \\
\midrule
\multirow{6}{*}{\textbf{Object-Related (3,493)}} & $\rightarrow$ Object Hallucination & 2,810 \\
 & $\rightarrow$ Object Existence/Count & 683 \\
 & $\rightarrow$ Visual Challenge (Complex) & 699 \\
 & $\rightarrow$ Scene Hallucination & 250 \\
 & $\rightarrow$ Action Hallucination & 123 \\
\midrule
\multirow{4}{*}{\textbf{Attribute-Related (1,622)}} & $\rightarrow$ Attribute Hallucination (State) & 1,169 \\
 & $\rightarrow$ Attribute Hallucination (Number) & 280 \\
 & $\rightarrow$ Attribute Hallucination & 163 \\
 & $\rightarrow$ Attribute Hallucination (Color) & 10 \\
\midrule
\textbf{Relationship (1,720)} & $\rightarrow$ Relationship Hallucination & 1,720 \\
\midrule
\multirow{6}{*}{\textbf{Other (3,165)}} & $\rightarrow$ General VQA & 670 \\
 & $\rightarrow$ Reasoning/Calculation & 407 \\
 & $\rightarrow$ Unanswerable/Insufficient Context & 355 \\
 & $\rightarrow$ Identity Hallucination & 340 \\
 & $\rightarrow$ Temporal Hallucination & 170 \\
 & $\rightarrow$ Text/OCR Hallucination & 151 \\
\bottomrule
\textbf{Total} & & \textbf{10,000} \\
\bottomrule
\end{tabular}%
}
\caption{Hallucination type distribution with grouped main categories and specific error subcategories.}
\label{tab:hallucination_distribution}
\end{table}

The hallucination taxonomy as seen in Table~\ref{tab:hallucination_distribution} demonstrates a comprehensive coverage of error patterns encountered in VLM outputs. \textbf{Object-Related} errors dominate with 3,493 instances, reflecting the fundamental challenge of object hallucination in vision-language models where fabricated objects (2,810 cases) represent the most common failure mode. The substantial representation of \textbf{Relationship} hallucinations (1,720 cases) highlights the difficulty models face in accurately describing spatial and logical relationships between visual elements. \textbf{Attribute-Related} errors (1,622 instances) predominantly involve state misattributions (1,169 cases), indicating models struggle with dynamic properties rather than static characteristics like color. The \textbf{Other} category encompasses diverse failure modes including reasoning errors and contextual misunderstandings, ensuring comprehensive evaluation coverage.

\paragraph{Answer Type Distribution}

The answer format distribution as showcased in Table~\ref{tab:answer_distribution} reflects a pragmatic balance between evaluation complexity and real-world applicability. The predominance of \textbf{Yes/No} questions (6,568 samples, 65.7\%) provides clear ground truth for binary classification while enabling efficient large-scale evaluation. \textbf{Open-Ended} responses (2,008 samples, 20.1\%) evaluate free-form generation capabilities and complex reasoning, representing scenarios where models must provide detailed explanations. The inclusion of \textbf{Unanswerable} questions (731 samples, 7.3\%) tests models' ability to recognize limitations and avoid overconfident responses—a critical capability for reducing hallucinations. \textbf{Numeric} responses (657 samples, 6.6\%) focus on quantitative reasoning precision, while \textbf{Selection} tasks (36 samples, 0.4\%) provide targeted evaluation of spatial and referential understanding.

This carefully composed dataset enables robust evaluation of hallucination prediction across the full spectrum of VLM capabilities, ensuring that \name's performance metrics reflect real-world deployment scenarios and diverse error patterns encountered in multimodal reasoning tasks.

\begin{table}[t]
\centering
\resizebox{\columnwidth}{!}{%
\begin{tabular}{l l r}
\toprule
\textbf{Answer Category} & \textbf{Definition} & \textbf{Count} \\
\midrule
\multirow{2}{*}{\textbf{Yes/No}} & $\rightarrow$ Binary answers: yes, no & \multirow{2}{*}{6,568} \\
 & $\rightarrow$ Simple true/false responses & \\
\midrule
\multirow{2}{*}{\textbf{Open-Ended}} & $\rightarrow$ Descriptive text answers & \multirow{2}{*}{2,008} \\
 & $\rightarrow$ Detailed explanations & \\
\midrule
{\textbf{Unanswerable}} & $\rightarrow$ "Can't tell"/"Don't know"  & {731} \\
\midrule
\multirow{2}{*}{\textbf{Number}} & $\rightarrow$ Numeric answers, counts & \multirow{2}{*}{657} \\
 & $\rightarrow$ Quantitative responses & \\
\midrule
\multirow{2}{*}{\textbf{Selection}} & $\rightarrow$ Multiple choice options & \multirow{2}{*}{36} \\
 & $\rightarrow$ "The one on the right", etc. & \\
\bottomrule
\textbf{Total} & & \textbf{10,000} \\
\bottomrule
\end{tabular}%
}
\caption{Answer type distribution showing response categories and their specific characteristics.}
\label{tab:answer_distribution}
\end{table}

\section{Data Construction}

We assemble the 10,000 sample evaluation set by sampling from six established VQA benchmarks. Figure~\ref{fig:Dataset}(a)–(c) summarize its composition:

\begin{enumerate}
  \item Domains (Figure~\ref{fig:Dataset}(a)): Attribute Recognition (30.1\%), Visual Understanding (29.8\%), Spatial Reasoning (17.7\%), Knowledge \& Identity (6.5\%), Math \& Calculation (6.3\%), Text \& OCR (5.1\%), General QA (2.7\%), and Temporal \& Video (1.7\%).
  \item Answer Formats (Figure~\ref{fig:Dataset}(b)): Yes/No (65.7\%), Open-Ended (20.1\%), Unanswerable (7.3\%), Numeric (6.6\%), and Selection (0.4\%).
  \item Hallucination Categories (Figure~\ref{fig:Dataset}(c)): Object-related (34.9\%), Other errors (31.7\%), Relationship-based (17.2\%), and Attribute-related (16.2\%).
\end{enumerate}

These distributions ensure broad coverage of visual and reasoning challenges. We then employ GPT-4 as an LLM judge \citep{zheng2023judging} with lenient criteria: flagging hallucinations only when the model invents nonexistent entities, contradicts the ground truth, or provides entirely incorrect information, but not for accurate paraphrases or added detail. The resulting labels $b^j = \mathrm{LLM}_{\mathrm{judge}}(\hat{Y}^j, Y^j, Q^j)$ serve as targets for our probes (Section~\ref{sec:llm-judge-eval}).

\section{Model-Specific Hallucination Patterns}

\medskip
\noindent

Table~\ref{tab:hallucination_rates_horizontal} reports the percentage of hallucinated outputs in four error categories: Attribute-Related, Object-Related, Relationship, and Other for each of eight Vision–Language Models. Lower percentages represent stronger resistance to that error mode. Notably, Qwen2.5-VL-7B achieves the lowest attribute-related (4.87 \%) and relationship (0.87 \%) rates, while FastVLM-7B minimizes object-related hallucinations (5.81 \%). The “Other” category is best constrained by Qwen2.5-VL-7B (9.13 \%). These results highlight each model’s relative strengths and weaknesses in pre-generation hallucination detection.

\section{LLM as a Judge}
\label{app:llm-judge}
To obtain reliable hallucination labels, we employ GPT-4 as the judge model \citep{zheng2023judging}. Formally,
\[
  b^j = \mathrm{LLM}_{\mathrm{judge}}(\hat{Y}^j, Y^j, Q^j),
\]
where the judge compares the model-generated response \(\hat{Y}^j\) against the ground truth \(Y^j\) in the context of query \(Q^j\). We use a lenient flagging strategy:

\begin{itemize}
  \item Flag as hallucinated (\(b^j = 1\)) only if:
    \begin{itemize}
      \item The response mentions entities/objects not present in the image.
      \item The response contradicts the ground truth factually.
      \item The response provides completely incorrect information.
    \end{itemize}
  \item Do \emph{not} flag (\(b^j = 0\)) if:
    \begin{itemize}
      \item The response paraphrases the ground truth accurately.
      \item The response adds correct details beyond the ground truth.
      \item The response uses different wording but conveys the same meaning.
    \end{itemize}
\end{itemize}

The prompt template given to GPT-4 is:

\begin{verbatim}
You are an expert judge for vision-language 
tasks.
Compare the generated answer: 
"<generated_answer>"
against the ground truth: 
"<ground_truth>"
for the question: "<question>".

Use the following lenient criteria:
- Flag as hallucination (output "1") 
only if the model invents nonexistent 
objects, contradicts the ground truth, 
or provides completely incorrect 
information.
- Otherwise, output "0".

Return only "0" or "1".
\end{verbatim}

\section{Comparative Analysis: Gemma-3-12B vs. FastVLM-7B}
Comparing Gemma-3-12B and FastVLM-7B (refer Figure \ref{fig:query-token-auroc}) reveals contrasting architectural strategies for encoding hallucination signals. Gemma-3-12B demonstrates the canonical progression pattern, with query tokens showing dramatic improvement from Layer 1 (0.7165) to Layer $L$ (0.9349), while vision tokens actually decline from 0.6538 to 0.5956. This suggests that Gemma-3-12B concentrates hallucination detection capabilities in late-stage multimodal reasoning while sacrificing pure visual signal strength.

FastVLM-7B exhibits the opposite behavior: vision tokens maintain relatively strong performance (0.6935 → 0.7028) with optimal performance at Layer $L/2$ (0.7093), while query tokens show minimal improvement and even slight decline (0.6715 → 0.6136). This architectural divergence indicates that FastVLM-7B preserves visual hallucination signals throughout processing rather than concentrating them in final reasoning states, possibly reflecting different training objectives or fusion strategies.

 \section{Threshold Analysis for Early Refusal}                                                                        
  \label{appendix:threshold}                                                                                            
\begin{table*}[t]                                                                                       
  \centering                                                                                                            
  \small                                                                                                                
  \begin{tabular}{ll|cccccccc}                                                                                          
  \toprule                                                                                                              
  \textbf{Metric} & \textbf{Rep} & \textbf{Qwen2.5} & \textbf{Gemma3} & \textbf{LLaVA} & \textbf{Molmo} &               
  \textbf{Llama3.2} & \textbf{Phi4} & \textbf{FastVLM} & \textbf{SmolVLM} \\                                            
  \midrule                                                                                                              
  \multirow{3}{*}{AUROC}                                                                                                
   & VF & 0.787 & 0.674 & 0.611 & 0.683 & 0.770 & 0.617 & 0.683 & 0.724 \\                                              
   & VT & 0.668 & 0.596 & 0.627 & 0.687 & 0.738 & 0.774 & 0.703 & 0.689 \\                                              
   & QT & 0.915 & 0.935 & 0.903 & 0.919 & 0.896 & 0.903 & 0.614 & 0.901 \\                                              
  \midrule                                                                                                              
  \multirow{3}{*}{Threshold}                                                                                            
   & VF & 0.200 & 0.100 & 0.200 & 0.100 & 0.300 & 0.100 & 0.300 & 0.100 \\                                                              
   & VT & 0.200 & 0.200 & 0.100 & 0.200 & 0.100 & 0.300 & 0.100 & 0.100 \\                                                              
   & QT & 0.200 & 0.200 & 0.200 & 0.100 & 0.100 & 0.200 & 0.100 & 0.400 \\                                                              
  \midrule                                                                                                              
  \multirow{3}{*}{Precision}                                                                                            
   & VF & 0.188 & 0.179 & 0.210 & 0.208 & 0.406 & 0.148 & 0.252 & 0.198 \\                                              
   & VT & 0.212 & 0.198 & 0.197 & 0.222 & 0.211 & 0.356 & 0.147 & 0.206 \\                                              
   & QT & 0.486 & 0.573 & 0.539 & 0.542 & 0.518 & 0.514 & 0.153 & 0.498 \\                                              
  \midrule                                                                                                              
  \multirow{3}{*}{Recall}                                                                                               
   & VF & 0.380 & 0.406 & 0.419 & 0.548 & 0.290 & 0.479 & 0.305 & 0.561 \\                                              
   & VT & 0.273 & 0.226 & 0.390 & 0.486 & 0.425 & 0.479 & 0.655 & 0.566 \\                                              
   & QT & 0.570 & 0.708 & 0.688 & 0.801 & 0.624 & 0.668 & 0.253 & 0.692 \\                                              
  \midrule                                                                                                              
  \multirow{3}{*}{F1}                                                                                                   
   & VF & 0.251 & 0.249 & 0.279 & 0.301 & 0.339 & 0.226 & 0.276 & 0.292 \\                                              
   & VT & 0.238 & 0.212 & 0.262 & 0.304 & 0.282 & 0.408 & 0.240 & 0.302 \\                                              
   & QT & 0.525 & 0.633 & 0.604 & 0.647 & 0.566 & 0.581 & 0.191 & 0.579 \\                                              
  \bottomrule                                                                                                           
  \end{tabular}                                                                                                         
  \caption{Complete early refusal analysis at optimal F1 threshold across all models and representation types. VT and QT
   use layer $L$.}                                                                                                      
  \label{tab:threshold_summary_full}                                                                                    
  \end{table*}

  Section~\ref{sec:control} introduced early refusal as a practical application of {\name} scores. Here we provide quantitative threshold
   analysis across all eight VLMs for early refusal, using the final decoder layer ($L$) for Vision Token (VT) and Query Token (QT)        
  representations. For each trained probe, we sweep classification thresholds $\tau \in \{0.1, 0.2, \ldots, 0.9\}$ and  
  identify the threshold that maximizes F1 score on the held-out test set.                                              
       
  \subsection{Query Token (QT) Performance}                                                                             
             
  Table~\ref{tab:qt_threshold} shows the optimal F1 thresholds for QT representations. Seven of eight models achieve    
  AUROC greater than 0.89 (Table~\ref{tab:vlm_comparison}), enabling early refusal of 57\% to 80\% of hallucinations with 48\% to 57\% precision at optimal  
  thresholds ranging from 0.1 to 0.4.                                                                                   
       
  \begin{table}[t]                                                                                                      
  \centering                                                            
  \resizebox{\columnwidth}{!}{
  \begin{tabular}{lccc}                                                                                                 
  \toprule                                                                                                              
  \textbf{Model} & \textbf{Best F1} & \textbf{Recall} & \textbf{Precision} \\                                           
   & \textbf{Threshold} & & \\                                                                                          
  \midrule                                                                                                              
  Gemma3-12B     & 0.2 & 70.8\% & 57.2\% \\                                                                             
  Molmo-7B       & 0.1 & 80.1\% & 54.2\% \\                                                                             
  LLaVA-1.5-7B   & 0.2 & 68.8\% & 53.9\% \\                                                                             
  SmolVLM-2.2B   & 0.4 & 69.2\% & 49.8\% \\                                                                             
  Phi4-VL        & 0.2 & 66.8\% & 51.4\% \\                                                                             
  Llama-3.2-11B  & 0.1 & 62.4\% & 51.8\% \\                                                                             
  Qwen2.5-VL-7B  & 0.2 & 57.0\% & 48.6\% \\                                                                             
  FastVLM        & 0.1 & 25.3\% & 15.3\% \\                                                                             
  \bottomrule                                                                                                           
  \end{tabular}                                                         }                             
  \caption{Last Query Token (QT) early refusal performance at optimal F1 threshold.}                                         
  \label{tab:qt_threshold}                                                                                              
  \end{table}                                                                                                           
                                                                                                                        
  \subsection{Vision Feature (VF) Performance}                                                                         
                                                                                                                        
  Table~\ref{tab:vf_threshold} demonstrates that even pure vision encoder features, before any LLM decoder processing, encode   
  hallucination-predictive signals. While precision is lower than QT, this enables the earliest possible intervention   
  point in the VLM pipeline.                                                                                            
                                                                                                                        
  \begin{table}[h]                                                                                                      
  \centering                                                            
  \resizebox{\columnwidth}{!}{
  \begin{tabular}{lccc}                                                                                                 
  \toprule                                                                                                              
  \textbf{Model} & \textbf{Best F1} & \textbf{Recall} & \textbf{Precision} \\                                           
   & \textbf{Threshold} & & \\                                                                                          
  \midrule                                                                                                              
  Qwen2.5-VL-7B  & 0.2 & 38.0\% & 18.8\% \\                                                                             
  Llama-3.2-11B  & 0.3 & 29.0\% & 40.6\% \\                                                                             
  SmolVLM-2.2B   & 0.1 & 56.1\% & 19.8\% \\                                                                             
  Molmo-7B       & 0.1 & 54.8\% & 20.8\% \\                                                                             
  FastVLM        & 0.3 & 30.5\% & 25.2\% \\                                                                             
  Gemma3-12B     & 0.1 & 40.6\% & 17.9\% \\                                                                             
  Phi4-VL        & 0.1 & 47.9\% & 14.8\% \\                                                                             
  LLaVA-1.5-7B   & 0.2 & 41.9\% & 21.0\% \\                                                                             
  \bottomrule                                                                                                           
  \end{tabular}     
  }
  \caption{Vision Feature (VF) (before LLM decoder) early refusal performance at optimal F1 threshold.}                                     
  \label{tab:vf_threshold}                                                                                              
  \end{table}                                                                                                           
       
  \subsection{Vision Token (VT) Performance} 

    Table~\ref{tab:vt_threshold} shows VT performance at layer L. Vision tokens enable detection of 23\% to 66\% of hallucinations with precision between 15\% and 36\%, with Phi4-VL achieving the strongest VT performance. 
    
\begin{table}[h]                                                                                                      
  \centering                                                            \resizebox{\columnwidth}{!}{                                 
  \begin{tabular}{lccc}                                                                                                 
  \toprule                                                                                                              
  \textbf{Model} & \textbf{Best F1} & \textbf{Recall} & \textbf{Precision} \\                                           
   & \textbf{Threshold} & & \\                                                                                          
  \midrule                                                                                                              
  Phi4-VL        & 0.3 & 47.9\% & 35.5\% \\                                                                             
  Llama-3.2-11B  & 0.1 & 42.5\% & 21.1\% \\                                                                             
  FastVLM        & 0.1 & 65.5\% & 14.7\% \\                                                                             
  SmolVLM-2.2B   & 0.1 & 56.6\% & 20.6\% \\                                                                             
  Molmo-7B       & 0.2 & 48.5\% & 22.2\% \\                                                                             
  Qwen2.5-VL-7B  & 0.2 & 27.3\% & 21.1\% \\                                                                             
  LLaVA-1.5-7B   & 0.1 & 39.0\% & 19.7\% \\                                                                             
  Gemma3-12B     & 0.2 & 22.6\% & 19.8\% \\                                                                             
  \bottomrule                                                                                                           
  \end{tabular}                                                         
  }
  \caption{Last Vision Token (VT) early refusal performance at optimal F1 threshold (layer $L$).}                            
  \label{tab:vt_threshold}                                                                                              
  \end{table}

  \subsection{Complete Summary}                                                                                         
             
  Table~\ref{tab:threshold_summary_full} provides the complete threshold analysis with all metrics (AUROC, optimal      
  threshold, precision, recall, and F1) across all models and representation types.    
 
  \subsection{Deployment Recommendations}                                     
  Based on the threshold analysis for early refusal for potential hallucination risks, we provide the following deployment guidelines:        
  \begin{itemize}                                                                                                       
      \item \textbf{Safety-critical applications}: Use QT probes with threshold $\tau$ between 0.1 and 0.2 to maximize            
  hallucination coverage. At these thresholds, models like Molmo-7B can catch up to 80\% of hallucinations before any   
  token is generated.                                                                                                   
      \item \textbf{Balanced deployment}: Use QT probes with $\tau$ between 0.2 and 0.4 for optimal precision-recall    
  trade-off, achieving 48\% to 57\% precision while maintaining 57\% to 71\% recall.                               
  
      \item \textbf{Earliest intervention}: Use VF probes when minimal latency is critical. Qwen2.5-VL-7B (AUROC 0.787) 
  and Llama-3.2-11B (AUROC 0.770) achieve the strongest VF performance, enabling hallucination risk assessment using    
  only visual encoder outputs.                                                                                          
  \end{itemize}                                                                                                         
       
  These results demonstrate that {\name} scores can serve as practical control signals for early refusal policies, trading 
  response coverage for higher reliability without requiring any text generation.                                                                   
\end{document}